\documentclass[11pt]{article}

\usepackage{amsmath, amssymb, amsfonts}
\usepackage{graphicx}
\usepackage{booktabs}
\usepackage[hidelinks]{hyperref}
\usepackage{caption}
\usepackage{subcaption}
\usepackage{placeins}
\usepackage{geometry}
\usepackage{float}

\usepackage{authblk}
\title{\textbf{IRAM-$\boldsymbol{\Omega}$-Q}:\\
A Computational Framework for Uncertainty Regulation in Adaptive Agents}

\author[1]{Veronique Ziegler}
\affil[1]{Independent Researcher}
\date{}

\begin{document}
\maketitle

\begin{abstract}
Adaptive agents operating under uncertainty must do more than optimize task outputs: they must maintain a workable internal state under noise, perturbation, and changing conditions. This paper introduces \textbf{IRAM-$\boldsymbol{\Omega}$-Q}, a computational framework for modeling uncertainty regulation in adaptive agents under stochastic disturbance. The framework combines a quantum-like state representation with closed-loop adaptive control over an internal entropy signal. The quantum-like formalism is used instrumentally: the evolving state is a normalized complex amplitude vector, coherent evolution is propagated exactly as $\psi(t+\Delta t)=e^{-iH\Delta t}\psi(t)$, and a derived density matrix supports entropy and coherence-gap analysis. Two causal control orderings are compared. In regulation-first (RF) ordering, adaptive regulation is available before current-cycle disturbance and attenuates incoming exposure; in disturbance-first (DF) ordering, current-cycle disturbance is received before a new regulatory response can be computed, and stabilization acts reactively. Publication-mode, matched-seed simulations show broadly comparable coherence-gap trajectories but lower sustained adaptive gain under RF. Susceptibility maps based on post--burn-in temporal fluctuations further show that DF shifts the critical initial-gain ridge toward larger values across multiple disturbance intervals. These results identify ordering as an architectural determinant of regulatory demand and threshold location within an otherwise shared regime structure.
\end{abstract}

\section{Introduction}

Adaptive agents operating under uncertainty must do more than optimize task outputs. They must also maintain a workable internal state in the presence of noise, perturbation, and changing conditions. A system that cannot regulate its internal uncertainty may become unstable, excessively diffuse, or overly rigid even when its external objective is well defined. This motivates computational frameworks in which internal regulation is treated as an explicit dynamical process rather than an implicit by-product of optimization.

Many existing approaches expose only part of this picture. Symbolic and explicitly structured models provide interpretability, but often do not directly model how internal uncertainty is dynamically regulated under persistent disturbance. End-to-end systems can be robust in practice, but typically offer limited visibility into how internal stability is maintained, when control demand rises, or where the system becomes transition-sensitive. A useful framework should make these internal regulatory properties measurable.

This paper introduces IRAM-$\Omega$-Q (Integrated Regulation and Attention Model with $\Omega$-regulation and a quantum-like state representation), a computational framework for uncertainty regulation in adaptive agents. The symbol $\Omega$ denotes the explicit adaptive regulation component of the architecture. Our goal is not to claim that biological or artificial cognition literally depends on microscopic quantum computation. Instead, we use a quantum-like formalism instrumentally, as a mathematically structured state representation that supports the joint analysis of uncertainty, control effort, state organization, and regime transitions \cite{busemeyer2012quantum}. Relative to most quantum-cognition models, which are usually descriptive and open-loop, IRAM-$\Omega$-Q is formulated as a closed-loop regulatory architecture. Relative to classical adaptive control, the regulated object is not simply an external tracking error, but the structured uncertainty of an internal state representation.

The present analysis asks three questions. First, can adaptive control maintain bounded internal uncertainty under stochastic disturbance? Second, do initial regulation and incoming disturbance jointly define reproducible transition-sensitive regions in parameter space? Third, does causal timing matter: is anticipatory regulation more efficient than reactive recovery when disturbance arrives before protective action can be established?

\paragraph{Contribution Summary}
This work makes three primary contributions.
\begin{itemize}
\item It introduces a closed-loop uncertainty-regulation mechanism within a quantum-like state model, with exact unitary coherent propagation and adaptive regulation driven by a derived entropy signal.
\item It identifies reproducible regime structure using a susceptibility statistic based on post--burn-in temporal variance of the coherence gap, and estimates an operational critical initial-gain boundary $\widehat{\mu}_{0,c}(\eta)$.
\item It compares two causally distinct orderings: regulation-first (RF), in which current-cycle disturbance is attenuated anticipatorily, and disturbance-first (DF), in which disturbance enters before a new regulatory response is computed. In publication-mode simulations, RF requires lower sustained adaptive gain, while DF shifts the susceptibility ridge toward higher initial regulation gain over multiple disturbance intervals.
\end{itemize}

The contribution is computational rather than metaphysical. IRAM-$\Omega$-Q provides operational definitions and measurable quantities aligned with uncertainty-aware AI, adaptive dynamical systems, and cognitively inspired agent modeling. It makes no claims regarding phenomenological consciousness or physical quantum processes.

\section{Conceptual Framework}

\subsection{What this framework is and is not}

IRAM-$\Omega$-Q is a computational framework for studying internal regulation dynamics in adaptive agents. It uses a quantum-like state representation because a normalized complex amplitude state provides a compact way to represent structured internal organization, while the associated density-matrix representation supports entropy and coherence-gap observables. The contribution of the framework is computational and analytical: it provides a closed-loop model in which internal uncertainty, regulatory effort, and regime structure can be measured from the same evolving state.

The model is not proposed as a claim that biological cognition depends on microscopic quantum processes. The quantum-like formalism is used as a state-space language for structured uncertainty and order effects. This choice is useful here because the paper asks how an internal state changes under coherent evolution, disturbance, and feedback regulation, and because the relevant observables can be defined directly from the same state representation.

\subsection{State representation, update loop, and analysis targets}

IRAM-$\Omega$-Q is formulated as a closed-loop dynamical system. At each time step, the agent carries an internal amplitude state; disturbance perturbs that state; an adaptive gain modulates stabilization and, when regulation is available before exposure, reduces incoming disturbance; exact coherent internal dynamics then propagate the state. The model is designed to answer a regulation question: \emph{can an adaptive controller maintain an internal state within a workable uncertainty regime despite ongoing disturbance, and how does this depend on whether regulation precedes or follows exposure?}

\paragraph{Primary dynamical state.}
The primary dynamical state is a normalized complex amplitude vector
\begin{equation}
\psi(t)=\bigl(\psi_1(t),\dots,\psi_d(t)\bigr)\in\mathbb{C}^d,
\qquad
\sum_{i=1}^{d}|\psi_i(t)|^2=1.
\end{equation}
Pure-state initialization uses a salience profile over the basis indices, after which the state is normalized. The simulation evolves $\psi(t)$ directly. For metric computation, the current amplitude state is mapped to a real-valued density-matrix representation
\begin{equation}
\rho_{ij}(t)=\Re\!\bigl(\psi_i(t)\overline{\psi_j(t)}\bigr),
\end{equation}
followed by trace normalization. Thus $\psi(t)$ is the primary evolving state, while $\rho(t)$ is the derived analysis state used to compute uncertainty and coherence-gap observables.

\paragraph{Exact coherent internal evolution.}
The coherent part of the dynamics is generated by a Hamiltonian $H$,
\begin{equation}
\dot{\psi}(t)=-\,iH\psi(t),
\end{equation}
and is propagated in the implementation by the exact finite-time unitary map
\begin{equation}
\psi(t+\Delta t)=U_{H,\Delta t}\psi_{\mathrm{nc}}(t),
\qquad
U_{H,\Delta t}=e^{-iH\Delta t},
\label{eq:unitary}
\end{equation}
where $\psi_{\mathrm{nc}}(t)$ denotes the state after the non-coherent disturbance and stabilization operations for the current cycle. The real symmetric Hamiltonian is diagonalized once for the fixed $(H,\Delta t)$ combination, and the resulting propagator is reused across time steps. The Hamiltonian has diagonal terms encoding intrinsic basis energies and off-diagonal couplings that decay exponentially with basis distance:
\begin{equation}
H_{ii}=E_i, \qquad
H_{ij}=c\,e^{-|i-j|/\ell}\quad(i\neq j),
\end{equation}
where $E_i$ is the energy associated with basis index $i$, $c$ is the coupling strength, and $\ell$ is the interaction length scale.

\paragraph{Disturbance and attentional stabilization.}
Let $\eta$ denote the incoming disturbance amplitude presented to the system during a cycle. A stabilization operation $\mathcal{M}_{\mu}$ acts when a focus index $f$ has been specified: it damps all non-focus amplitudes by $(1-\mu)$, leaves the focus component unchanged, and then renormalizes,
\begin{equation}
\psi_i \leftarrow \psi_i \quad \text{if } i=f,
\qquad
\psi_i \leftarrow (1-\mu)\psi_i \quad \text{if } i\neq f.
\end{equation}
This operation does not directly amplify the focus component; it suppresses competing components so that the focus becomes relatively more dominant after normalization.

The realized disturbance exposure depends on causal ordering. When regulation acts before exposure, the incoming disturbance is attenuated by the updated gain,
\begin{equation}
\eta_{\mathrm{eff}}^{\mathrm{RF}}(t)
=
\eta\bigl(1-\mu_{\mathrm{RF}}^{+}(t)\bigr).
\label{eq:eta-rf}
\end{equation}
When disturbance arrives before the new regulatory response, the current-cycle exposure is not retroactively attenuated,
\begin{equation}
\eta_{\mathrm{eff}}^{\mathrm{DF}}(t)=\eta.
\label{eq:eta-df}
\end{equation}
This difference is not an unequal-parameter comparison: both orderings are presented with the same incoming disturbance amplitude $\eta$. It is the modeled consequence of whether protective regulation is available before that disturbance reaches the state.

\paragraph{Adaptive regulation gain.}
The scalar gain $\mu(t)$ is the controller output and the model's direct control-effort variable. The controller is driven by the von Neumann entropy of the derived state and by a finite-difference estimate of entropy change,
\begin{equation}
\frac{dS}{dt}\approx S_{\mathrm{vN}}(t)-S_{\mathrm{vN}}(t-\Delta t).
\end{equation}
With base gains $\alpha_0$ and $\beta_0$, target entropy $S^\ast$, and learning-scale factor $\gamma$, the scheduled gains are
\begin{equation}
\alpha(t)=\gamma\,\alpha_0\bigl(1-\mu(t)\bigr),
\qquad
\beta(t)=\gamma\,\beta_0\,\mu(t),
\end{equation}
and the gain update is
\begin{equation}
\mu(t+\Delta t)=
\Pi_{[\mu_{\min},\mu_{\max}]}
\left[
\mu(t)+\alpha(t)\frac{dS}{dt}
+\beta(t)\bigl(S_{\mathrm{vN}}(t)-S^\ast\bigr)
\right],
\label{eq:mu-update}
\end{equation}
where $\Pi_{[\mu_{\min},\mu_{\max}]}$ denotes clipping to the admissible gain interval. With the error convention $S_{\mathrm{vN}}(t)-S^\ast$, entropy above target gives a positive contribution through the $+\beta(t)$ term and therefore increases regulatory gain, as required by the feedback interpretation. Because $\mu(t)$ evolves during a trajectory, the phase sweep varies the \emph{initial} regulation gain $\mu_0$, not a fixed gain held constant throughout the run.

\paragraph{Regulation-first and disturbance-first orderings.}
The two conditions represent anticipatory protection versus reactive recovery. Let $\mathcal{D}_{\eta}$ denote the stochastic disturbance operation and let $\mathcal{C}$ denote the entropy-driven controller update.

In regulation-first (RF) ordering, the controller observes the pre-disturbance state, obtains an updated gain $\mu_{\mathrm{RF}}^{+}(t)$, stabilizes the state, and attenuates incoming disturbance before exposure:
\begin{align}
\mu_{\mathrm{RF}}^{+}(t)
&=
\mathcal{C}\!\left(\rho(t),\mu(t)\right),\\
\psi_{\mathrm{nc}}^{\mathrm{RF}}(t)
&=
\mathcal{D}_{\eta(1-\mu_{\mathrm{RF}}^{+}(t))}
\!\left[
\mathcal{M}_{\mu_{\mathrm{RF}}^{+}(t)}\!\left(\psi(t)\right)
\right],\\
\psi^{\mathrm{RF}}(t+\Delta t)
&=
U_{H,\Delta t}\psi_{\mathrm{nc}}^{\mathrm{RF}}(t).
\end{align}

In disturbance-first (DF) ordering, the incoming disturbance reaches the state before a new protective update is available. The controller subsequently observes the disturbed state, updates the gain, and applies reactive stabilization:
\begin{align}
\widetilde{\psi}^{\mathrm{DF}}(t)
&=
\mathcal{D}_{\eta}\!\left(\psi(t)\right),\\
\mu_{\mathrm{DF}}^{+}(t)
&=
\mathcal{C}\!\left(\widetilde{\rho}^{\mathrm{DF}}(t),\mu(t)\right),\\
\psi_{\mathrm{nc}}^{\mathrm{DF}}(t)
&=
\mathcal{M}_{\mu_{\mathrm{DF}}^{+}(t)}
\!\left(\widetilde{\psi}^{\mathrm{DF}}(t)\right),\\
\psi^{\mathrm{DF}}(t+\Delta t)
&=
U_{H,\Delta t}\psi_{\mathrm{nc}}^{\mathrm{DF}}(t).
\end{align}
The ordering comparison therefore asks whether protection available before incoming disturbance reduces control demand and shifts transition-sensitive thresholds relative to recovery that begins only after exposure.

\begin{figure*}[ht]
\centering
\includegraphics[width=1.0\textwidth]{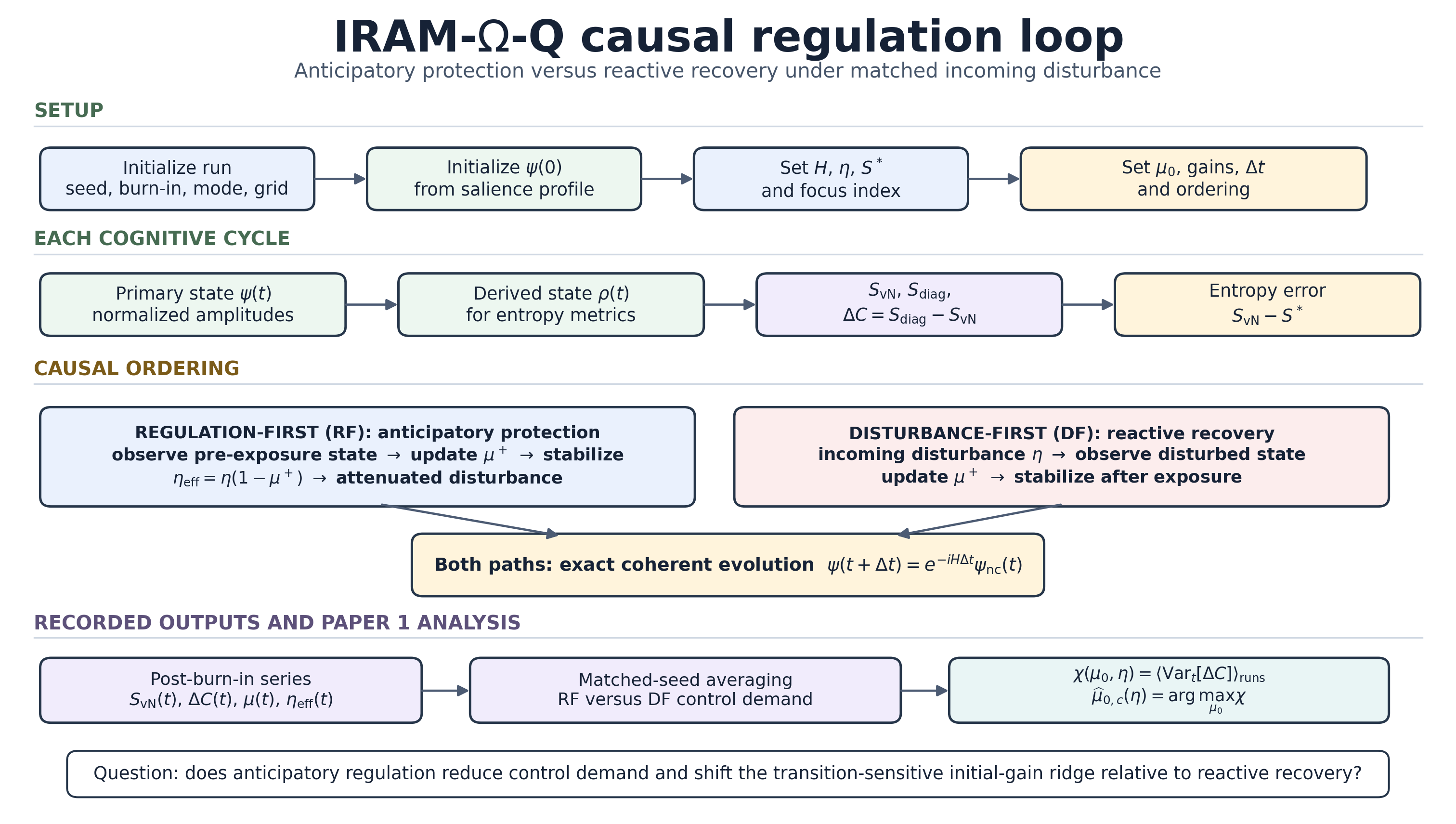}
\caption{\textbf{IRAM-$\Omega$-Q closed-loop update.}
The primary state is a normalized amplitude vector $\psi(t)$; a derived metric state $\rho(t)$ provides the entropy signal $S_{\mathrm{vN}}$ that drives adaptive regulation gain $\mu(t)$. Regulation-first (RF) computes a new gain before exposure, applies stabilization, and attenuates the current incoming disturbance. Disturbance-first (DF) receives the incoming disturbance before a new gain is computed and applies stabilization reactively. Both paths end with the same exact coherent propagator $U_{H,\Delta t}=e^{-iH\Delta t}$. Post--burn-in trajectories are summarized through $S_{\mathrm{vN}}(t)$, $\Delta C(t)=S_{\mathrm{diag}}(t)-S_{\mathrm{vN}}(t)$, $\mu(t)$, temporal susceptibility $\chi(\mu_0,\eta)$, and the operational boundary $\widehat{\mu}_{0,c}(\eta)$.}
\label{fig:loop_overview}
\end{figure*}
\FloatBarrier

\section{Metrics and Observables}
\label{sec:metrics}

The framework tracks complementary observables: the uncertainty signal used by the controller, the amount of adaptive gain required, the structural organization retained by the state, and the temporal fluctuation structure that reveals transition-sensitive operating regions.

\paragraph{Von Neumann entropy.}
The controlled uncertainty variable is the von Neumann entropy of the derived density matrix,
\begin{equation}
S_{\mathrm{vN}}(\rho)=-\mathrm{Tr}(\rho\log\rho)
=-\sum_{k=1}^{d}\lambda_k\log\lambda_k,
\end{equation}
where $\{\lambda_k\}$ are the eigenvalues of $\rho$. This is the entropy signal used by the controller.

\paragraph{Diagonal cognitive entropy.}
To distinguish full-state uncertainty from basis-population uncertainty, we compute the diagonal entropy
\begin{equation}
S_{\mathrm{diag}}(\rho)=-\sum_{i=1}^{d}\rho_{ii}\log \rho_{ii}.
\end{equation}
This treats the diagonal entries of $\rho$ as a classicalized distribution over basis populations.

\paragraph{Coherence gap.}
The principal state-structure observable in the sweep analysis is the coherence gap
\begin{equation}
\Delta C(\rho)=S_{\mathrm{diag}}(\rho)-S_{\mathrm{vN}}(\rho).
\end{equation}
Larger $\Delta C$ indicates greater organization in the full-state representation relative to its diagonalized comparison. For each initial-gain and incoming-disturbance condition $(\mu_0,\eta)$, the post--burn-in temporal mean in run $r$ is
\begin{equation}
\overline{\Delta C}_{r}(\mu_0,\eta)=
\frac{1}{|\mathcal{T}_{\mathrm{post}}|}
\sum_{t\in\mathcal{T}_{\mathrm{post}}}
\Delta C_r(t;\mu_0,\eta).
\end{equation}

\paragraph{Regulation gain and incremental update.}
The scalar variable $\mu(t)$ is the adaptive controller output and therefore the direct measure of regulatory demand. Its incremental update is
\begin{equation}
\Delta\mu(t)=\mu(t)-\mu(t-\Delta t).
\end{equation}
In matched-seed comparisons, lower sustained $\mu(t)$ accompanied by comparable state observables indicates that an ordering achieves similar regulation with lower control effort.

\paragraph{Temporal susceptibility and critical initial-gain boundary.}
Mean coherence-gap values alone do not identify transition-sensitive dynamics. The publication sweep therefore uses temporal fluctuation susceptibility after burn-in:
\begin{equation}
\chi(\mu_0,\eta)
=
\frac{1}{R}
\sum_{r=1}^{R}
\mathrm{Var}_{t\in\mathcal{T}_{\mathrm{post}}}
\!\left[
\Delta C_r(t;\mu_0,\eta)
\right],
\label{eq:susceptibility}
\end{equation}
where $R$ is the number of matched-protocol stochastic replicates. Large $\chi$ identifies conditions for which coherence-gap dynamics fluctuate strongly over time after transients are removed. Because $\mu(t)$ remains adaptive during the sweep, the horizontal control parameter is the initial gain $\mu_0$. The corresponding operational boundary is
\begin{equation}
\widehat{\mu}_{0,c}(\eta)
=
\arg\max_{\mu_0}
\chi(\mu_0,\eta).
\label{eq:critical-initial}
\end{equation}
This estimator identifies the sampled initial gain associated with maximal post--burn-in temporal susceptibility at each incoming disturbance amplitude. It is an operational marker of transition sensitivity, not a mathematically exact bifurcation point or guaranteed minimum-safe gain.

\paragraph{Summary of roles.}
$S_{\mathrm{vN}}$ is the regulated uncertainty signal; $\mu(t)$ is regulatory demand; $S_{\mathrm{diag}}$ provides the diagonal comparison; $\Delta C$ characterizes full-state organization relative to that comparison; $\chi(\mu_0,\eta)$ maps post--burn-in temporal sensitivity; and $\widehat{\mu}_{0,c}(\eta)$ summarizes the location of its ridge.

\section{Experimental Methods}

All reported results were produced in publication mode using deterministic base seeds, matched RF/DF seed mixing, explicit burn-in windows, and automatically exported numerical outputs. Exploratory low-resolution runs were used only to validate the pipeline before launching the publication-resolution sweep.

\subsection{Simulation Protocol}

Two causal orderings were evaluated under identical incoming disturbance schedules and matched stochastic seeds:
\begin{itemize}
\item \textbf{Regulation first (RF):} adaptive regulation and stabilization are established before incoming disturbance, so current-cycle exposure is attenuated according to Eq.~\eqref{eq:eta-rf}.
\item \textbf{Disturbance first (DF):} incoming disturbance acts before a new regulatory response is computed, so current-cycle exposure is unattenuated according to Eq.~\eqref{eq:eta-df}, followed by reactive stabilization.
\end{itemize}
The difference in realized exposure is therefore the causal consequence of timing, not a difference in the externally specified incoming disturbance amplitude $\eta$.

\subsection{Matched-Seed Trajectory Comparison}

For the trajectory comparison, RF and DF were run with identical initial settings and the same replicate seed sequence. Each condition used $30$ stochastic replicates with $\mu_0=0.08$, $\eta=0.13$, target entropy $S^\ast=0.30$, $15{,}000$ integration steps, and a recorded-sample burn-in of $1{,}000$ samples. The figures report across-run means and one-standard-deviation bands for $\Delta C(t)$ and $\mu(t)$.

\subsection{Publication-Resolution Susceptibility Sweep}
\label{sec:phase-methods}

For each ordering, the phase analysis swept initial regulation gain and incoming disturbance amplitude over
\begin{equation}
\mu_0\in[0.001,0.50]\quad(50\ \text{grid values}),
\qquad
\eta\in[0.001,0.30]\quad(40\ \text{grid values}).
\end{equation}
At each grid point, $20$ stochastic replicate trajectories were generated using matched seed construction across RF and DF. Each trajectory comprised $15{,}000$ steps, with the first $5{,}000$ steps discarded before computing the temporal susceptibility in Eq.~\eqref{eq:susceptibility}. The RF and DF susceptibility maps are plotted with a shared color normalization so that identical colors represent identical susceptibility values across orderings.

\subsection{Operational Critical Initial-Gain Boundary}
\label{sec:critical}

For each fixed incoming disturbance amplitude $\eta$, the operational boundary $\widehat{\mu}_{0,c}(\eta)$ is the sampled initial gain that maximizes temporal susceptibility, as defined in Eq.~\eqref{eq:critical-initial}. Because gain adapts after initialization, this curve must be interpreted as a \emph{critical initial-gain ridge}, not as the response of a system with fixed $\mu$.

\begin{table}[H]
\centering
\caption{\textbf{Publication-mode simulation parameters used for the updated Paper~1 figures.}}
\label{tab:simulation-params}
\begin{tabular}{lll}
\toprule
\textbf{Analysis} & \textbf{Parameter} & \textbf{Value} \\
\midrule
All runs & State dimension / focus index & $16 / 6$ \\
         & $\Delta t$ & $0.01$ \\
         & Salience center / width & $6.0 / 2.0$ \\
         & Phase noise & $0.2$ \\
         & Energy scale / coupling / locality & $0.15 / 0.08 / 2.0$ \\
         & Target entropy $S^\ast$ & $0.30$ \\
         & Controller gains $(\alpha_0,\beta_0)$ & $(5\times10^{-4},\,2\times10^{-4})$ \\
         & Gain bounds $(\mu_{\min},\mu_{\max})$ & $(0.001,\,1.0)$ \\
         & Base seed & $123456789$ \\
\midrule
Trajectories & Orderings & RF and DF, matched seeds \\
             & $(\mu_0,\eta)$ & $(0.08,\,0.13)$ \\
             & Steps / replicates & $15{,}000 / 30$ \\
             & Burn-in samples & $1{,}000$ \\
\midrule
Susceptibility maps & Orderings & RF and DF, matched seeds \\
                    & $\mu_0$ grid & $[0.001,0.50]$, $50$ values \\
                    & $\eta$ grid & $[0.001,0.30]$, $40$ values \\
                    & Steps / burn-in & $15{,}000 / 5{,}000$ \\
                    & Runs per grid point & $20$ \\
\bottomrule
\end{tabular}
\end{table}

\FloatBarrier

\section{Results}
\label{sec:results}

\subsection{Matched-seed trajectory comparison: comparable state dynamics with reduced RF control demand}

\begin{figure}[htbp]
\centering
\begin{subfigure}{0.98\linewidth}
  \centering
  \includegraphics[width=\linewidth]{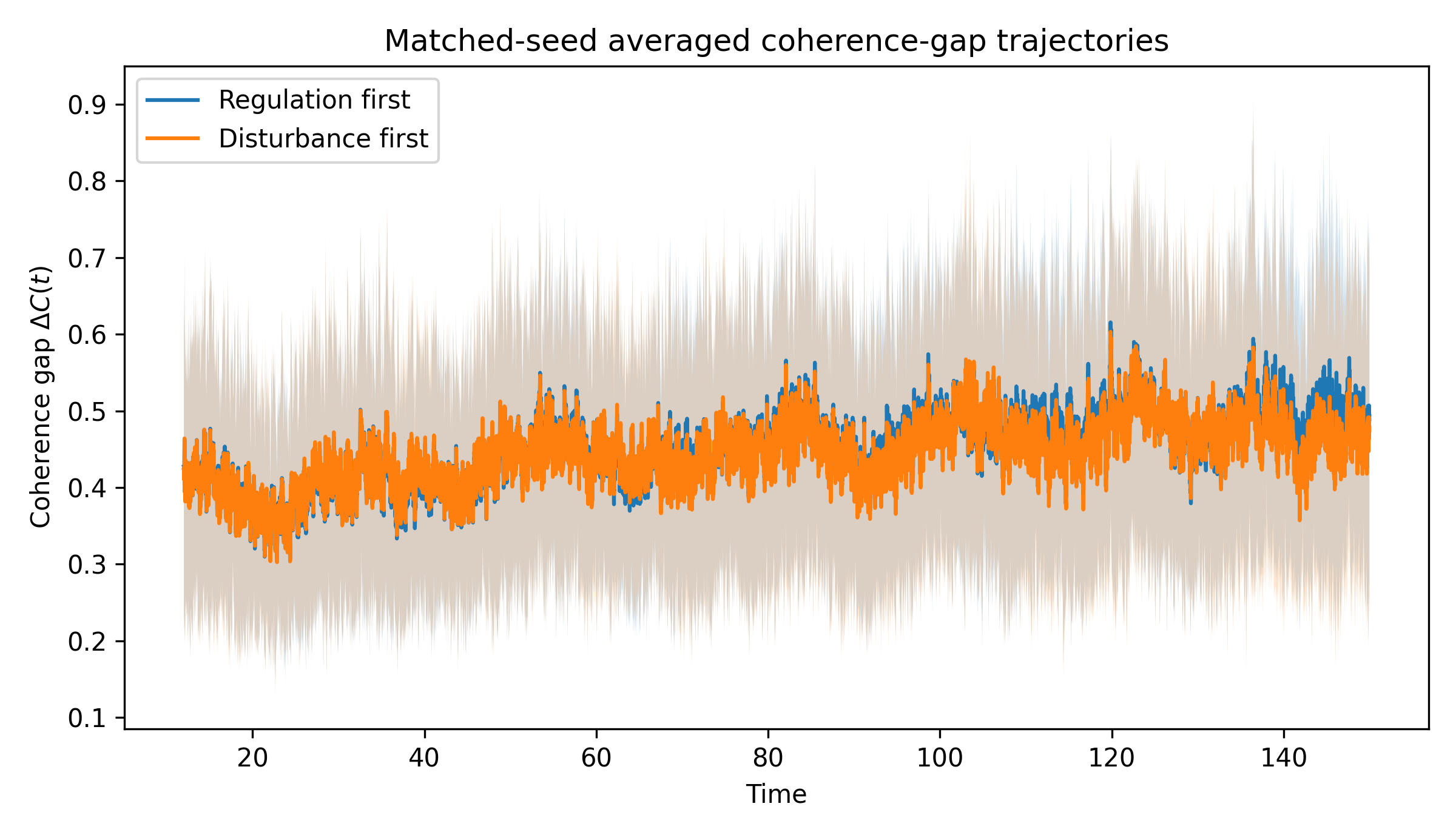}
  \caption{Matched-seed averaged coherence-gap trajectories.}
\end{subfigure}

\vspace{0.8em}

\begin{subfigure}{0.98\linewidth}
  \centering
  \includegraphics[width=\linewidth]{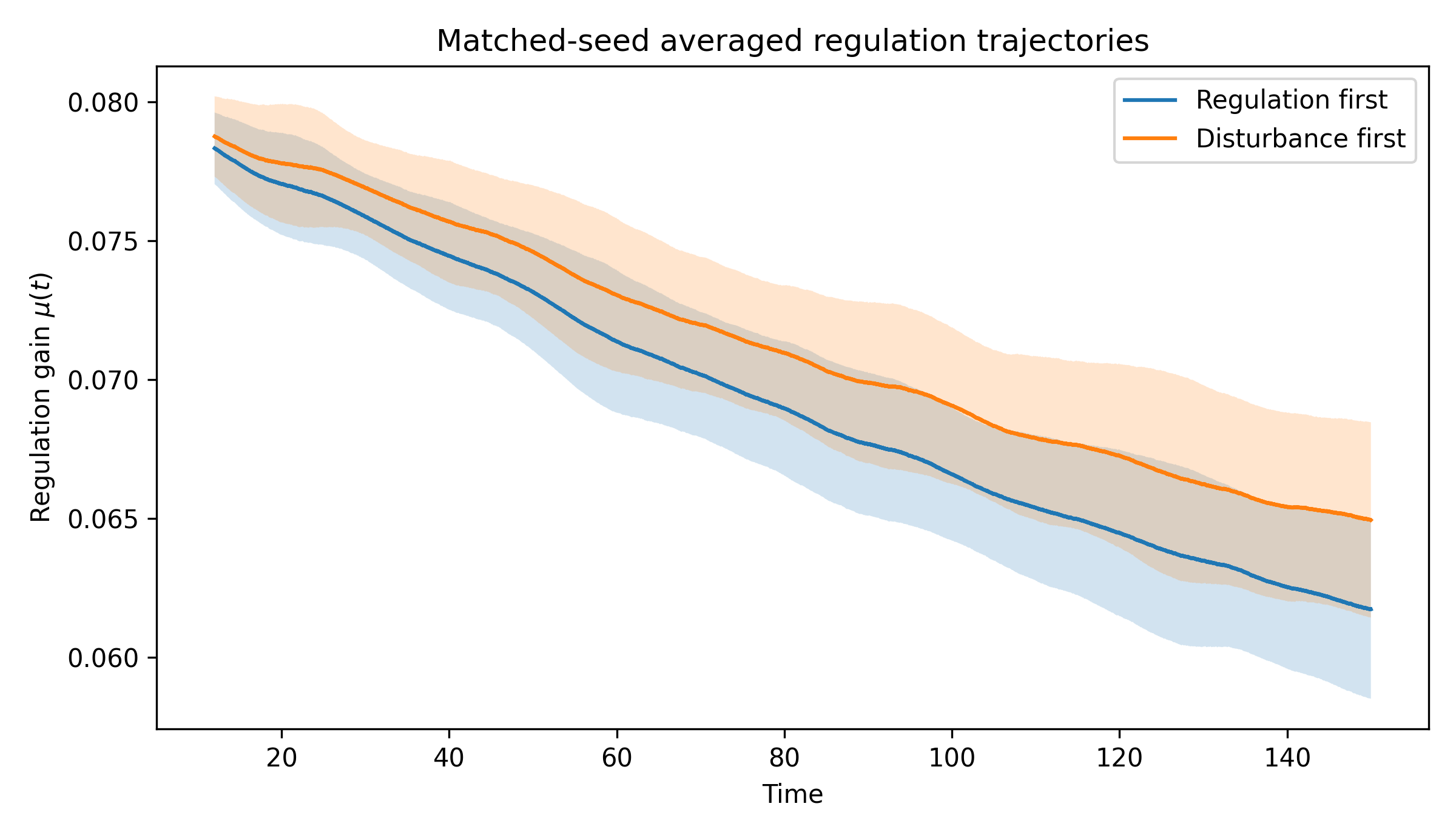}
  \caption{Matched-seed averaged adaptive regulation-gain trajectories.}
\end{subfigure}
\caption{\textbf{Trajectory-level comparison of causal control ordering.}
Regulation first (RF) and disturbance first (DF) are compared with identical incoming disturbance amplitude $\eta=0.13$, initial gain $\mu_0=0.08$, target entropy $S^\ast=0.30$, and matched stochastic seeds. Lines show across-run means and shaded regions show $\pm 1\sigma$ over $30$ replicates. The coherence-gap traces are broadly similar, while the lower sustained $\mu(t)$ under RF indicates reduced regulatory demand relative to reactive DF.}
\label{fig:trajectory-ordering}
\end{figure}

Figure~\ref{fig:trajectory-ordering} compares RF and DF trajectories under identical incoming disturbance amplitude, initial gain, target entropy, and replicate seeds. The mean coherence-gap trajectories are broadly comparable and their uncertainty bands strongly overlap (Fig.~\ref{fig:trajectory-ordering}a). Thus the causal ordering change does not generate a wholesale reorganization of the state-structure observable in this operating condition.

The regulation-gain trajectories reveal the principal difference (Fig.~\ref{fig:trajectory-ordering}b). After the early transient, the RF mean gain remains below the DF mean gain through most of the recorded interval. RF therefore attains comparable coherence-gap behavior while maintaining lower adaptive regulatory demand. In the model's causal interpretation, anticipatory regulation is more efficient because it is available to attenuate incoming disturbance before exposure, whereas DF must recover after unattenuated current-cycle disturbance has already entered.

\subsection{Temporal susceptibility identifies a low-initial-gain transition ridge}

\begin{figure}[htbp]
\centering
\begin{subfigure}{0.49\linewidth}
  \centering
  \includegraphics[width=\linewidth]{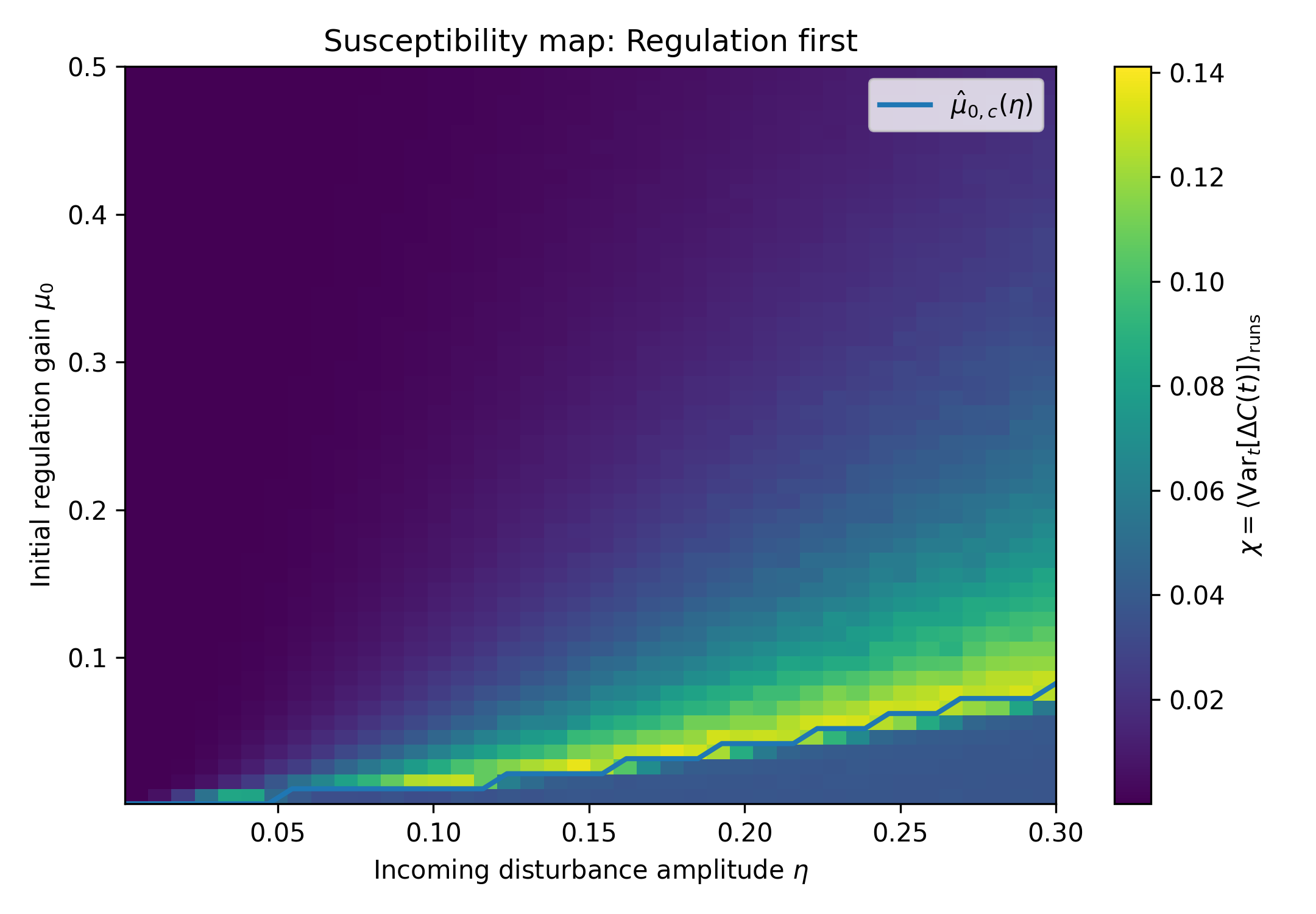}
  \caption{Regulation first (RF).}
\end{subfigure}\hfill
\begin{subfigure}{0.49\linewidth}
  \centering
  \includegraphics[width=\linewidth]{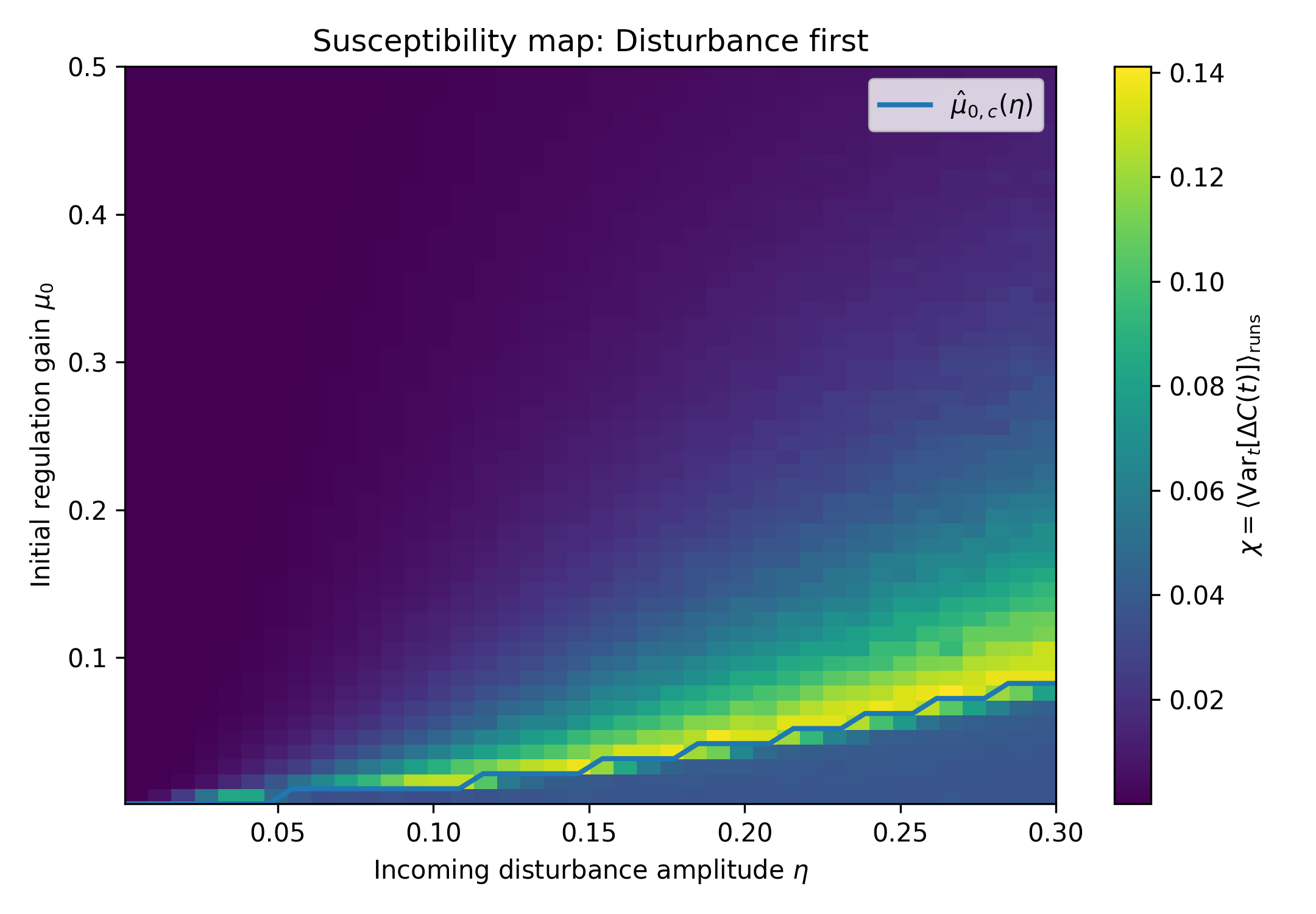}
  \caption{Disturbance first (DF).}
\end{subfigure}
\caption{\textbf{Publication-resolution susceptibility maps under causal ordering.}
Temporal susceptibility is $\chi(\mu_0,\eta)=\langle \mathrm{Var}_t[\Delta C(t)]\rangle_{\mathrm{runs}}$ after burn-in. The vertical axis is the initial adaptive gain $\mu_0$, since $\mu(t)$ continues to evolve during each trajectory. The overlaid line is the operational critical initial-gain ridge $\widehat{\mu}_{0,c}(\eta)$. A shared color scale permits direct comparison of susceptibility magnitudes across RF and DF.}
\label{fig:susceptibility-ordering}
\end{figure}

Figure~\ref{fig:susceptibility-ordering} presents publication-resolution temporal susceptibility maps for both causal orderings. For each grid point, susceptibility is computed as the replicate average of post--burn-in temporal variance in $\Delta C(t)$, Eq.~\eqref{eq:susceptibility}. Both maps use a common color scale and show the same broad topology: a transition-sensitive ridge is concentrated near the low-$\mu_0$ boundary and rises as incoming disturbance amplitude increases.

The overall similarity is expected because RF and DF share the same state space, Hamiltonian, controller law, and incoming-disturbance grid. The ordering effect appears not as an entirely different landscape, but as a displacement in the susceptibility ridge. At several disturbance intervals, the DF ridge lies at a larger initial regulation gain than the RF ridge.

\subsection{Disturbance-first ordering increases critical initial-gain demand in resolved intervals}

\begin{figure}[htbp]
\centering
\includegraphics[width=0.96\linewidth]{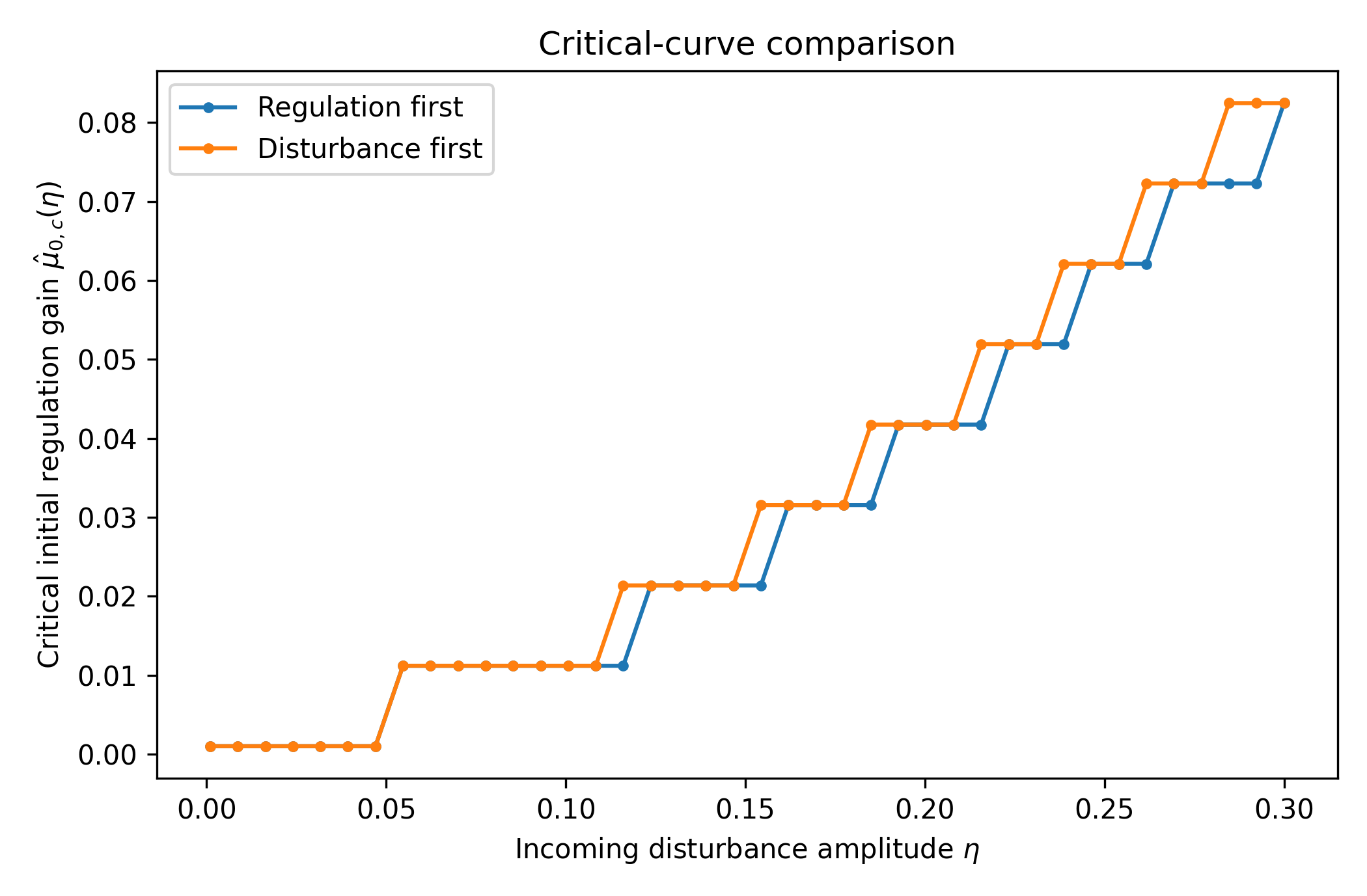}
\caption{\textbf{Direct comparison of operational critical initial-gain ridges.}
The susceptibility-derived ridge $\widehat{\mu}_{0,c}(\eta)$ is shown for RF and DF orderings. The stepped appearance reflects finite-grid maximization over the sampled $\mu_0$ values. DF is equal to or above RF throughout the resolved comparison and is shifted upward over multiple incoming-disturbance intervals, indicating increased initial regulatory demand under reactive ordering.}
\label{fig:critical-overlay}
\end{figure}

Figure~\ref{fig:critical-overlay} overlays the critical initial-gain ridges extracted from the susceptibility maps. The two curves coincide in some disturbance intervals, confirming that ordering does not produce a uniform global offset. In multiple resolved intervals, however, the DF curve steps upward before the RF curve:
\begin{equation}
\widehat{\mu}_{0,c}^{\mathrm{DF}}(\eta)
>
\widehat{\mu}_{0,c}^{\mathrm{RF}}(\eta).
\end{equation}
This pattern supports the intended architectural interpretation: when incoming disturbance precedes newly computed protection, the transition-sensitive boundary shifts toward greater required initial regulation. Conversely, RF can remain on the comparable critical-ridge structure with lower initial regulatory demand over those intervals.

\FloatBarrier

\section{Discussion}

IRAM-$\Omega$-Q is intended as a computational framework for studying internal regulation dynamics in adaptive agents. Its contribution is not a behavioral benchmark or a claim about consciousness, but a formal setting in which internal uncertainty, control demand, and transition-sensitive regime structure can be measured jointly. The updated implementation strengthens that contribution in two ways: coherent internal evolution is propagated exactly by $e^{-iH\Delta t}$, and the ordering comparison is formulated causally as anticipatory protection versus reactive recovery.

The trajectory comparison shows that ordering need not produce large differences in the coherence-gap signal to matter operationally. RF and DF maintain broadly comparable $\Delta C(t)$ trajectories at the tested operating point, yet RF maintains a lower adaptive gain $\mu(t)$ over most of the retained interval. Within the model, this constitutes an efficiency advantage: regulation that can act before exposure requires less sustained gain than regulation that begins after unattenuated disturbance has entered.

The observed RF/DF gain separation should not be interpreted as evidence that
anticipatory control requires a quantum-like substrate; rather, the
quantum-like state representation makes it possible to study how a general
causal-ordering advantage is expressed through coherence-sensitive dynamics,
susceptibility structure, and critical-ridge behavior.

The susceptibility analysis complements that trajectory-level result. Because the sweep retains adaptive $\mu(t)$ dynamics, it maps sensitivity as a function of the initial gain $\mu_0$ and incoming disturbance amplitude $\eta$. The resulting ridge $\widehat{\mu}_{0,c}(\eta)$ is not a fixed-control bifurcation curve; it is an operational initial-condition boundary under adaptive regulation. The publication-resolution results show that DF shifts this ridge upward over multiple disturbance intervals while leaving the broad landscape intact. Thus the central ordering effect is a structured increase in regulatory demand rather than wholesale alteration of the phase geometry.

The density-matrix formalism is used as a compact derived representation of the evolving amplitude state. Its value here is pragmatic: it allows the same evolving state to support a controller-facing uncertainty measure $S_{\mathrm{vN}}$, a diagonal comparison entropy $S_{\mathrm{diag}}$, and a coherence-gap observable $\Delta C=S_{\mathrm{diag}}-S_{\mathrm{vN}}$. The present results do not establish that such regulation or threshold effects are uniquely quantum-like; comparable effects may arise in classical adaptive systems. The claim is narrower: this representation provides a coherent computational language for structured internal-state regulation and its causal timing effects.

\subsection{Future plans and possible extensions}

The present paper establishes the core regulation loop, causal RF/DF comparison, and susceptibility-based initial-gain ridge. Natural next steps include embedding the internal regulator in explicit task environments, connecting $S_{\mathrm{vN}}$, $\Delta C$, and $\mu$ to externally measurable robustness; comparing the model to classical adaptive-control baselines under matched disturbance schedules; and extending scalar gain to context- or component-dependent regulation.

Path-dependent protocols are particularly important. Continuous hysteresis schedules, recovery after perturbation, and dwell-time analysis can ask whether a system that has experienced strong disturbance carries regulatory history into subsequent conditions. Those analyses are reserved for subsequent work rather than folded into the present paper claims.

\subsection{Limitations}

Several limitations should be noted. First, the present study evaluates internal-state dynamics rather than explicit behavioral performance. Second, the density-matrix representation is a modeling choice and does not imply microscopic quantum cognition. Third, the stochastic disturbance operator is a computational disturbance/decoherence-like operation rather than a claim of physical environmental decoherence. Fourth, $\widehat{\mu}_{0,c}(\eta)$ is a finite-grid, finite-run estimator of temporal susceptibility under adaptive gain dynamics, not a proven bifurcation or universal minimum-safe gain. Finally, the RF advantage is established for the tested parameter family and causal disturbance model; broader parameter sweeps and independent implementation checks are required before claiming universality.

\FloatBarrier

\appendix
\section*{Supplementary Appendix: Estimation and Reproducibility Details}
\addcontentsline{toc}{section}{Supplementary Appendix: Estimation and Reproducibility Details}

\subsection*{S1. Operational estimator for the critical initial-gain ridge $\widehat{\mu}_{0,c}(\eta)$}

For each fixed incoming disturbance amplitude $\eta$, initial regulation gain is swept over a discrete grid $\{\mu_{0,1},\ldots,\mu_{0,M}\}$. Gain remains adaptive during every trajectory; therefore $\mu_0$ denotes an initialization condition rather than a fixed control value. For replicate run $r$, define the post--burn-in temporal susceptibility
\begin{equation}
\chi_r(\mu_0,\eta)
=
\mathrm{Var}_{t\in\mathcal{T}_{\mathrm{post}}}
\!\left[
\Delta C_r(t;\mu_0,\eta)
\right].
\end{equation}
The reported susceptibility is the replicate average,
\begin{equation}
\widehat{\chi}(\mu_0,\eta)
=
\frac{1}{R}\sum_{r=1}^{R}\chi_r(\mu_0,\eta),
\end{equation}
and the operational ridge is
\begin{equation}
\widehat{\mu}_{0,c}(\eta)
=
\arg\max_{\mu_{0,k}}\,
\widehat{\chi}(\mu_{0,k},\eta).
\end{equation}
This estimator identifies the sampled initial regulation gain at which post--burn-in coherence-gap dynamics are maximally temporally fluctuation-sensitive for each incoming disturbance amplitude.

\subsection*{S2. Causal interpretation of realized disturbance}

Both RF and DF conditions are specified with the same incoming amplitude $\eta$. Their difference is architectural timing. RF establishes a newly computed protective gain before exposure and therefore applies $\eta(1-\mu_{\mathrm{RF}}^{+})$ during the current cycle. DF receives $\eta$ before a new gain can be computed and then stabilizes reactively. The resulting difference in realized current-cycle exposure is therefore a modeled outcome of anticipatory versus reactive regulation, not a difference in the externally assigned disturbance condition.

\subsection*{S3. Reproducibility}

All updated results and figures were generated from deterministic headless runs of the IRAM-$\Omega$-Q simulation codebase. Trajectory figures use matched RF/DF replicate seeds, $30$ replicates, $15{,}000$ steps, and a $1{,}000$-sample recorded burn-in. Susceptibility maps use matched RF/DF seed construction, a $50\times40$ $(\mu_0,\eta)$ grid, $20$ runs per grid point, $15{,}000$ steps per run, and a $5{,}000$-step burn-in. The RF and DF susceptibility figures use a shared color normalization to permit direct comparison of susceptibility magnitude.

\FloatBarrier

\end{document}